\documentclass[10pt,twocolumn,letterpaper]{article}

\usepackage{iccv}
\usepackage{times}
\usepackage{graphicx}
\usepackage{amsmath}
\usepackage{amssymb}
\usepackage{subfigure}
\usepackage{makecell}
\usepackage{multirow}
\usepackage{bbm}

\usepackage[T1]{fontenc}    
\usepackage{booktabs}       

\iccvfinalcopy 

\usepackage{amsmath,amsfonts,bm}

\def\eqref#1{equation~\ref{#1}}

\def\1{\bm{1}}

\def\rvp{{\mathbf{p}}}

\def\mI{{\bm{I}}}

\def\mM{{\bm{M}}}

\def\mP{{\bm{P}}}

\DeclareMathAlphabet{\mathsfit}{\encodingdefault}{\sfdefault}{m}{sl}
\SetMathAlphabet{\mathsfit}{bold}{\encodingdefault}{\sfdefault}{bx}{n}

\newcommand{\R}{\mathbb{R}}

\def\iccvPaperID{****} 

\ificcvfinal\fi

\begin{document}

\title{DPMix: Mixture of Depth and Point Cloud Video Experts \\ for 4D Action Segmentation}


\author{Yue Zhang\\
College of Computer Science and Technology\\
Zhejiang University\\
{\tt\small cheong.yue@hotmail.com}
\and
Hehe Fan\thanks{Corresponding author.}\\
College of Computer Science and Technology\\
Zhejiang University\\
{\tt\small hehefan@zju.edu.cn}
\and
Yi Yang~~~~~~~~~~~~\\
College of Computer Science and Technology~~~~~~~~~~~~\\
Zhejiang University~~~~~~~~~~~~\\
{\tt\small yangyics@zju.edu.cn}~~~~~~~~~~~~
\and
Mohan Kankanhalli~~~~~~~~~~\\
School of Computing~~~~~~~~~~\\
National University of Singapore~~~~~~~~~~\\
{\tt\small mohan@comp.nus.edu.sg}~~~~~~~~
}

\maketitle
\ificcvfinal\thispagestyle{empty}\fi

\begin{abstract}
In this technical report, we present our findings from the research conducted on the Human-Object Interaction 4D (HOI4D) dataset for egocentric action segmentation task.
As a relatively novel research area, point cloud video methods might not be good at temporal modeling, especially for long point cloud videos (\eg, 150 frames).
In contrast, traditional video understanding methods have been well developed. Their effectiveness on temporal modeling has been widely verified on many large scale video datasets. Therefore, we convert point cloud videos into depth videos and employ traditional video modeling methods to improve 4D action segmentation. By ensembling depth and point cloud video methods, the accuracy is significantly improved.
The proposed method, named \textbf{Mix}ture of \textbf{D}epth and \textbf{P}oint cloud video experts (DPMix), achieved the first place in the 4D Action Segmentation Track of the HOI4D Challenge 2023.
\end{abstract}

\section{Introduction}\label{sec:intro}
The HOI4D dataset \cite{liu2022hoi4d} is a large-scale 4D egocentric video dataset that includes a wide range of hand-object interaction activities in indoor scenes. 
The original HOI4D dataset consists of 2.4M RGB-D egocentric  frames of over 4000 video sequences.
The dataset is collected by 9 participants who interacted with 800 different object instances of 16 object categories.
The dataset is captured in 610 different indoor rooms.

4D Action Segmentation is one of the tasks based on the HOI4D dataset. 
The task aims to learn an action recognition model that predicts the action category label for each point cloud frame in a given point cloud video. 
For 4D action segmentation, the dataset includes 3863 point cloud videos, with 2971 videos for training and 892 videos for evaluation, and 19 action categories. 
Each video consists of 150 frames. 
Each frame has 2048 points with 3D coordinates. 
The coordinate of the center point of  each frame is also provided.

Due to the capability of directly describing movements in the 3D space, point cloud videos play a key role in comprehending environmental changes and supporting interactions with the world. 
However, learning from 3D point cloud videos and understanding the dynamic 3D world is significantly challenging. It requires effective modeling for both 3D spatial and 1D temporal structure, especially for long videos with 150 frames.

The current methods, based on directly modeling original point cloud videos \cite{liu2019meteornet, fan2021iclr, fan2021deep, fan2022point, fan2019pointrnn},  can  preserve the original geometric information and thus provide an effectively understanding of 3D shape and appearance.
For example, Point 4D Transformer (P4Transformer) \cite{fan2021point} utilizes Transformers to capture local appearance and motion information of the entire video to avoid point tracking.
On the basis of P4Transformer, Point Primitive Transformer (PPTr) \cite{wen2022point} leverages the primitive plane as mid-level representation to capture the long-term spatial-temporal context in 4D point cloud videos.
However, as a relatively new research area, those point-based modeling approaches might not be good enough at temporal modeling, especially for long point cloud videos.
In contrast, traditional video understanding methods have been well developed. Their effectiveness on temporal modeling have been widely verified on many large scale video datasets. 
However, the geometry information in 2D images is less rich and precise compared to the information provided by the coordinate positions of points in 3D point clouds.

To address the above limitations, we propose DPMix that ensembles point cloud video-based experts and depth video-based experts.
On the one hand, we employ P4Transformer and PPTr as point cloud video experts since they have been proved effective on point cloud video modeling.
On the other hand, we project 3D point cloud frames to 2D depth image sequences, and utilize traditional RGB video modeling methods, including C3D \cite{tran2015learning}, R(2+1)D \cite{tran2018closer} and SlowFast \cite{feichtenhofer2019slowfast}, as depth video experts.
Finally, average fusion is performed on the predictions of point cloud video experts and depth video experts for 4D action recognition. 
With the proposed method, we achieve the first place in the 4D Action Segmentation Track of the HOI4D Challenge 2023. 

\section{Our Method}
\begin{figure*}[t!]
\centering
\includegraphics[width=\textwidth]{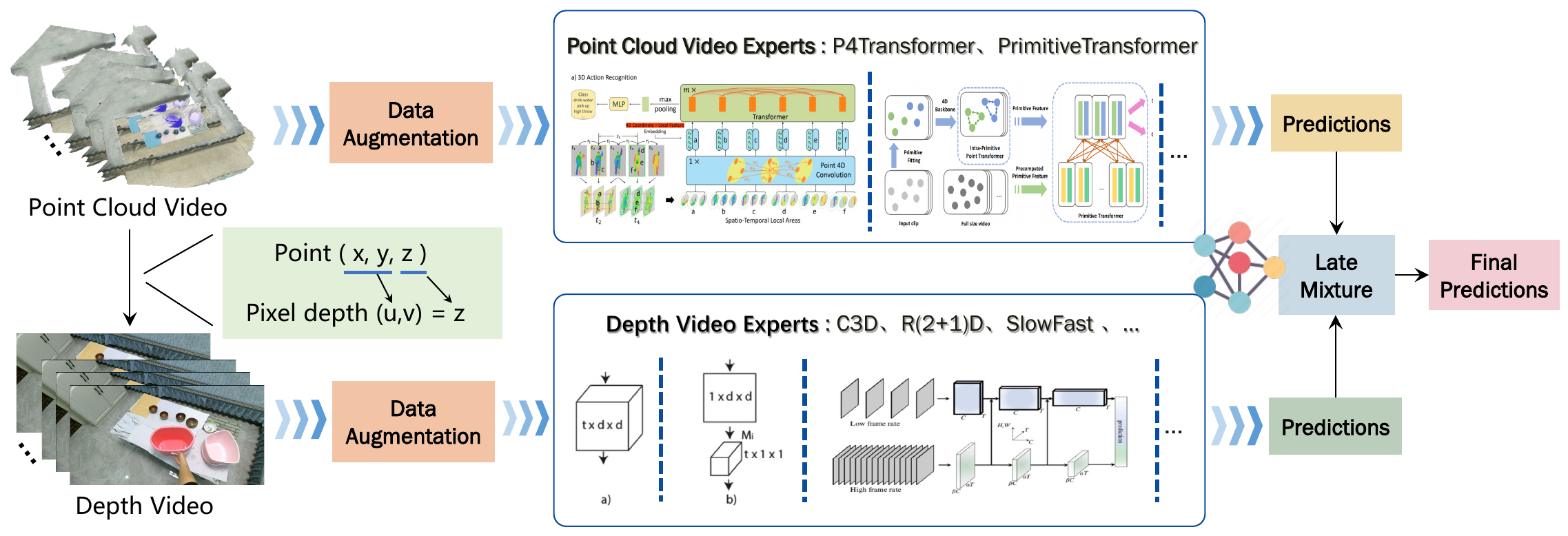}
\caption{Overall architecture of the proposed DPMix method. 
Point cloud video experts excel in capturing shape and appearance information using the geometric information of point clouds.
Deep video experts perform better temporal modeling, capturing long-range and short-range dependencies between hand motions.
The DPMix method is a late mixture of depth and point cloud video experts for 4D action segmentation.
}
\label{fig:architecture}
\end{figure*}

In this section, we elaborate on the technical details of our method.
As illustrated in Fig.~\ref{fig:architecture}, the overall architecture of DPMix consists two components: (a) the point cloud video experts and (b) the depth video experts. 
The DPMix ensembles depth and point cloud video experts for 4D action segmentation.
We will describe each component of the proposed method as follows.
\subsection{Point Cloud Video Modeling}
\textbf{Data Augmentation}.  
In order to enhance the size and quality of training dataset, we employ data augmentation including flipping, jittering, and scaling, on original input point cloud videos.
For each frame in a video, the point cloud is first moved to a non-offset position by subtracting the center point coordinates.
Then, we perform data transformation operations such as data scaling on the translated point cloud. 
Finally, the point cloud is shifted back to the original position by adding the center point coordinates.


\textbf{Point Cloud Video Experts}.
The point cloud video, a sequence of 3D coordinate collections combined with features, typically covers precise geometric information, which can provide greater flexibility for action recognition task.
The coordinate collection of each point cloud frame is unordered and unstructured, making neural-network-based modeling and processing extremely challenging.
We employ the popular P4Transformer network \cite{fan2021point} to model raw point cloud videos.
The P4Transformer designs a point four-dimensional convolution to embed the spatial-temporal local structure presented in a point cloud video, and then performs self-attention on the embedded local features through a Transformer \cite{vaswani2017attention} to capture the appearance and motion information of the entire video.

Long point cloud videos (\eg, 150 frames) pose challenges to existing Transformer-based backbone ( P4Transformer network \cite{fan2021point}) in terms of both effectiveness and efficiency.
On one hand, points between different frames are unstructured and inconsistent, thus effectively capturing the long-term spatial-temporal structure in long videos is difficult.
On the other hand, as the video length increases, 4D data is prone to memory and computational explosion.
For instance, a 24GB graphics card can only handle a synthia4D \cite{ros2016synthia} point cloud video of 3 frames \cite{wen2022point}.
To overcome above issues, research \cite{wen2022point} proposes PPTr. The PPTr extracts short-term spatial-temporal features through an intra-Primitive Point Transformer for a short video clip around interested frames, and extracts the long-term spatial-temporal features through a Primitive Transformer for the entire video clip. 
PPTr is proven to be efficient and effective in a wide range of 4D point cloud tasks, so we still employ this competitive 4D backbone network for the HOI4D action segmentation task.

\subsection{Depth Video Modeling}
\textbf{Convert to Depth Video}.
Point cloud video modeling methods can make full use of geometric information and have a better understanding of shape and appearance.
However, as a relatively new research area, those point-based modeling approaches might not be good enough at temporal modeling, especially for long point cloud videos.
By comparison, traditional video understanding methods have been well developed \cite{vahdani2022deep}. Their effectiveness on temporal modeling have been widely verified on many large scale video datasets. 
Based on above considerations, we transformed the input point cloud videos into depth videos and utilize the conventional video modeling methods.

Let $\mP_t \in \mathbb{R}^{N \times 3}$ denote the point coordinates of the $t$-th frame in a point cloud video, where $N$ defines the number of points and $3$ represents the three-dimensional coordinate representation of a point.
Hence, a point cloud video can be formalized as $\mathbf{\emph{PC}} = (\mP_1,  \mP_2, \cdots ,\mP_L) \in \R^{L \times N \times 3}$, where $L$ is the number of frames in video.
Eq.~\ref{eq:depth} formulates the method of converting point cloud videos to depth videos.
Specifically, for each point $\rvp \in \mP_t$ with coordinates $(x,y,z)$, we project the coordinates $(x, y)$ onto the pixel coordinates $(u, v)$ of an image. The coordinate $z$ of the point is regarded as the depth information of the corresponding pixel.
\begin{equation}
  \begin{aligned}
&u = (x - x_{min}) \times \frac{W}{(x_{max} - x_{min}) + \epsilon},\\
&v = (y - y_{min}) \times \frac{H}{(y_{max} - y_{min}) + \epsilon},\\
&depth(u, v) = z.
\end{aligned}  
\label{eq:depth}
\end{equation}

In the above equation, $W$ and $H$ are the width and height set for the converted depth image.
The $x_{max}$ and $x_{min}$ are the maximum and minimum values of the $x$ coordinates of all points in the point cloud video $\mathbf{\emph{PC}}$, respectively.
Similarly, $y_{max}$ and $y_{min}$ are the maximum and minimum values of the $y$ coordinates of all points in the point cloud video $\mathbf{\emph{PC}}$, respectively.
Besides, $depth(u, v)$ represents the depth value at the pixel with coordinates $(u, v)$ in the projected image.

After that, the $t$-th frame $\mP_t$ of a point cloud video $\mathbf{\emph{PC}}$ is converted into a depth image defined as $\mI_t \in \mathbb{R}^{W \times H \times C}$, where $W$, $H$, and $C$ are the width, height, and number of channels of the image, respectively.
A depth video can be formalized as $\mathbf{\emph{DV}} = (\mI_1,  \mI_2, \cdots ,\mI_L) \in \mathbb{R}^{L \times W \times H \times C}$, where $L$ represents the length of the depth video.
Note that since there is no RGB information in the original point cloud video, the converted video only contains single-channel depth information, \ie, $C = 1$.

\textbf{Data Augmentation}.  
We also adopt the data augmentation on the converted depth images to to increase the amount of training data and improve the generalization ability of models.
Region dropout strategies are effective for guiding convolutional neural network classifiers to attend on less discriminative parts of objects.
Nevertheless, the region dropout methods of removing pixels on images by overlaying a patch of either black pixels or random noise is not desirable, because these methods will lead to information loss in sparse depth images during training.
Although mixing parts of two images into a new image is a counterintuitive approach, it has proven to be a very effective augmentation strategy \cite{shorten2019survey}.
Therefore, the CutMix \cite{yun2019cutmix} augmentation strategy is exploited in this work.
The core idea of CutMix is to cut and paste patches among training images where the ground truth action labels are also mixed proportionally to the area of the patches.

\begin{table*}[t!]
    \centering\setlength{\tabcolsep}{15pt}\small
    \caption{Accuracy of different methods on each action category. The symbol ``Y" indicates that the model adopts the CutMix data augmentation strategy, while the symbol ``N" indicates that it does not.}
    \resizebox{2.1\columnwidth}{!}{
    \begin{tabular}{l|c|c|c|c|c|c|c|c}
        \toprule
       \multirow{2}*{Action}  & \multirow{2}*{P4Transformer}	& \multirow{2}*{PPTr}	& \multicolumn{2}{c|} {C3D}		& \multicolumn{2}{c|} {R(2+1)D}	& \multicolumn{2}{c} {SlowFast}	\\ 
       \cline{4-9} 
         & & & N & Y & N & Y & N & Y\\ \midrule
        Rest	& 89.39 	& 91.76 	& 93.66 	& 93.76 	& 93.81 	& 94.91 	& 94.18 	& 93.92 \\ \midrule
        Reachout	& 75.06 	& 83.97 	& 86.01 	& 86.42 	& 88.45 	& 88.45 	& 89.59 	& 87.10 \\ \midrule
        Grasp	& 29.08 	& 48.97 	& 59.15 	& 64.52 	& 68.38 	& 77.12 	& 70.56 	& 70.98 \\ \midrule
        Pickup	& 48.21 	& 60.07 	& 67.47 	& 69.05 	& 75.40 	& 72.07 	& 70.24 	& 75.79 \\ \midrule
        Carry	& 71.51 	& 77.72 	& 79.50 	& 83.09 	& 79.92 	& 84.24 	& 83.30 	& 81.27 \\ \midrule
        Putdown	& 57.17 	& 74.21 	& 79.05 	& 85.73 	& 84.62 	& 87.34 	& 88.16 	& 88.54 \\ \midrule
        Stop	& 68.85 	& 81.48 	& 83.79 	& 85.69 	& 82.35 	& 85.77 	& 86.84 	& 89.67 \\ \midrule
        Close	& 67.00 	& 74.07 	& 81.21 	& 75.36 	& 84.91 	& 71.60 	& 82.92 	& 83.40 \\ \midrule
        Dump	& 66.14 	& 76.09 	& 80.43 	& 78.15 	& 82.17 	& 81.30 	& 81.30 	& 79.57 \\ \midrule
        Open	& 71.17 	& 73.86 	& 82.23 	& 76.77 	& 82.11 	& 82.80 	& 80.14 	& 85.98 \\ \midrule
        Push	& 16.59 	& 61.82 	& 69.93 	& 60.61 	& 53.72 	& 48.65 	& 68.92 	& 53.72 \\ \midrule
        Pull	& 21.88 	& 54.92 	& 73.56 	& 64.74 	& 62.71 	& 76.27 	& 76.95 	& 73.56 \\ \midrule
        Papercut	& 49.33 	& 53.01 	& 79.52 	& 65.33 	& 68.07 	& 65.06 	& 59.64 	& 74.70 \\ \midrule
        Press	& 1.30 	& 14.47 	& 34.21 	& 42.21 	& 22.37 	& 43.42 	& 30.26 	& 30.26 \\ \midrule
        Carryboth	& 53.57 	& 62.69 	& 84.71 	& 89.96 	& 82.57 	& 80.43 	& 86.24 	& 85.32 \\ \midrule
        Binding	& 66.88 	& 79.52 	& 87.95 	& 88.96 	& 93.98 	& 96.39 	& 83.13 	& 97.59 \\ \midrule
        Cut	& 44.66 	& 68.10 	& 82.82 	& 80.24 	& 67.48 	& 65.64 	& 70.55 	& 60.74 \\ \midrule
        Turnon	& 32.70 	& 45.98 	& 80.46 	& 66.04 	& 71.26 	& 68.97 	& 66.67 	& 59.77 \\ \midrule
        Go	& 35.29 	& 30.88 	& 64.71 	& 72.06 	& 50.00 	& 55.88 	& 58.82 	& 57.35 \\ 
        \bottomrule
    \end{tabular}
    }
    \label{tab:val_results}
\end{table*}

Formally, let $\mI\in \R^{W \times H \times C}$ and $gt$ denote a frame and its label in a depth video, respectively.
The CutMix is designed to generate a new training sample $(\mI_{cm}, gt_{cm})$ by combining two training samples $(\mI_{t}, gt_{t})$ and $(\mI_{h}, gt_{h})$,
\begin{equation}
    \begin{aligned}
        & \mI_{cm} = \mM \odot \mI_t + (\mathbf{\emph{1}} - \mM) \odot \mI_h, \\
        & gt_{cm} = \lambda \cdot gt_{t} + (1 - \lambda) \cdot gt_{h},
    \end{aligned}
    \label{eq:cutmix}
\end{equation}
where $\mM \in \{0,1\}^{W \times H}$ represents a binary mask indicating the location of deletion and filling from two images, $\mathbf{\emph{1}}$ is a binary mask with dimension $W \times H$ filled with ones, and $\odot$ represents the Hadamard product operation.
The $\lambda$ denotes the combined ratio between the two images, which is sampled from the beta distribution Beta($\alpha, \alpha$).

\textbf{Depth Video Experts}.
To extract feature representations from the depth video frames, we leverage three popular models: 3D Convolutional Networks (C3D) \cite{tran2015learning}, R(2+1)D \cite{tran2018closer} and SlowFast \cite{feichtenhofer2019slowfast}.
C3D is a good feature learning machine that utilizes a deep 3D ConvNet to simultaneously model appearance and motion information.
R(2+1)D improves C3D by decomposing spatial and temporal modeling into two separate steps.
SlowFast adopts a slow path operating at low frame rate to capture spatial semantics, and a fast path operating at high frame rates to capture motion at fine temporal resolution.
The traditional video modeling methods we use above are all based on convolutional neural network architecture.
We speculate that video understanding models based on the Transformer architecture, such as Multiscale Vision Transformers (MViT) \cite{fan2021multiscale}, can also perform well in this task.

\subsection{DPMix}
Our DPMix method uses both depth video and point cloud video as input. On one hand, the point cloud video experts can acquire more accurate appearance information.
On the other hand, the depth video experts excel at capturing motion information.
The fusion of the two types of models can enhance each other in modeling the spatial and temporal dimensions.
For details, we perform the late collaboration by averaging the predictions from depth experts and point cloud experts, resulting in the improved accuracy in 4D action segmentation.

\section{Experiments}
In this section, we present the implementation details and give main experimental results and analysis.

\subsection{Implementation Details}
Following the challenge guidelines, We first split 90\% of the provided labeled data into training set and 10\% into validation set.
Then, our full models are trained on training set and evaluated on the validation set for algorithm validation and hyper-parameters tuning.
Finally, retrain and save the models on the complete training data with selected hyper-parameters.

Besides, implementation details of point cloud video experts (including P4Transformer and PPTr) and depth video experts (including C3D, R(2+1)D, and SlowFast) are described below.

\textbf{P4Transformer}.
The temporal radius is set as 3, the temporal stride is set as 1, the spatial radius is set as 0.9, the spatial sub-sampling rate is set as 32, 
, and the batch size is set as 64.
The Transformer contains 10 self-attention blocks, with 8 heads per block.
We train P4Transformer for 120 epochs using SGD optimizer and the initial learning rate is set as 0.05.

\textbf{PPTr}.
The spatial sub-sampling rate is set to 64 and the batch size is set to 16.
The other parameter settings are the same as P4Transformer.

\textbf{C3D}.
In converting point cloud video to depth video, we set the size of the input image to $112 \times 112$ for the C3D model.
When performing CutMix data augmentation, we set $\alpha$ in beta distribution $Beta(\alpha, \alpha)$ to $1$, that is $\lambda$ is sampled from the uniform distribution $(0,1)$.
We train C3D for 120 epochs using SGD optimizer with a batch size of 16, and the initial learning rate is set as 0.05.

\begin{figure*}[t!]
\centering
\includegraphics[width=\textwidth]{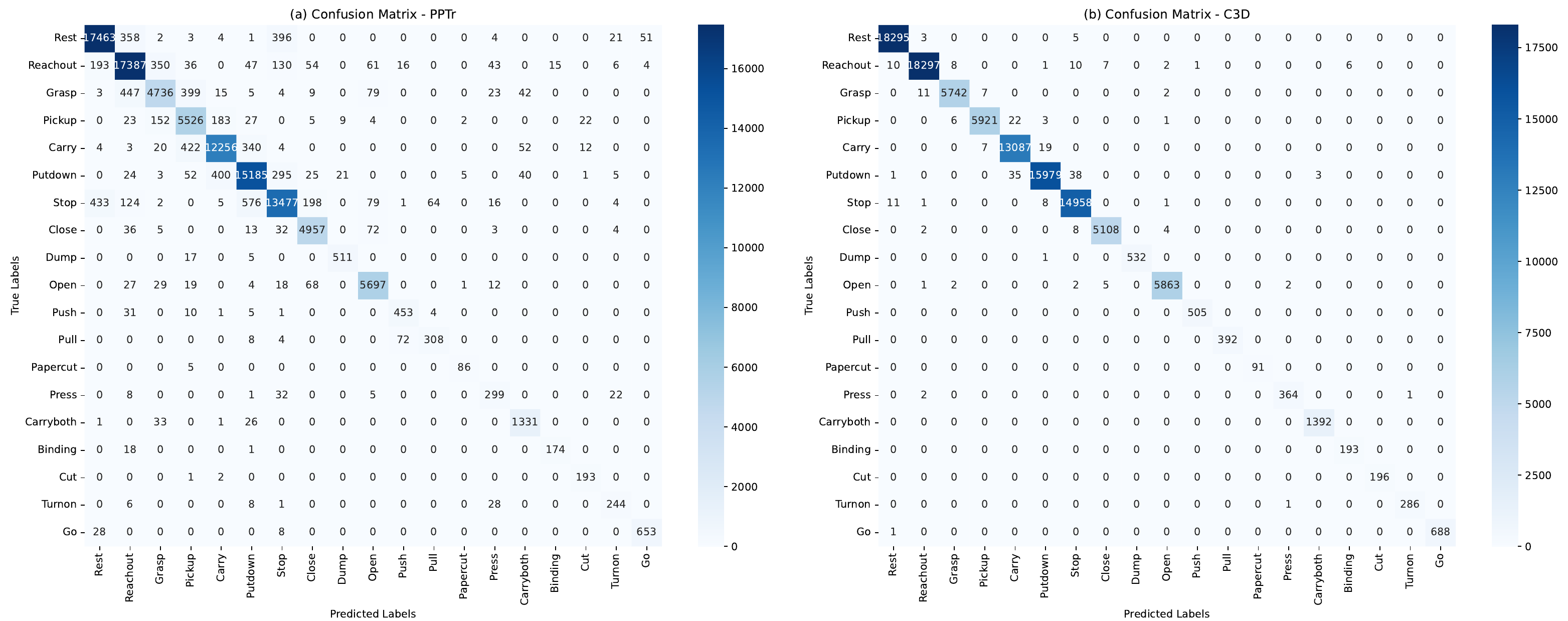}
\caption{Partial confusion matrices of point cloud video experts and depth video experts among 19 action categories on validation data. (a) is the confusion matrix of the PPTr. (b) is the confusion matrix of the C3D.}
\label{fig:confusion}
\end{figure*} 

\textbf{R(2+1)D}.
The input video dimensions, data augmentation parameter settings, and model training parameter settings are all the same as C3D.

\textbf{SlowFast}.
We set the size of the input image to $224 \times 224$ for the SlowFast with ResNet50 \cite{he2016deep}.
The number of frames is set as 150 and 32 for the fast and slow paths, respectively.
We train SlowFast for 200 epochs using SGD optimizer with a batch size of 8, and the initial learning rate is set as 0.01.

\subsection{Results and Analysis}
After training models on 90\% of the training data, we evaluate the accuracy on each action category on the remaining 10\% of the labeled data, and the results are shown in Table~\ref{tab:val_results}.
We observe that traditional video methods perform better at motion capture.
For action categories such as ``grasp", ``push", and ``press", where the accuracy of the P4Transformer is initially poorer, there is a significant improvement of depth video experts.
We also find that the CutMix data augmentation 
method can effectively improve the classification accuracy on action categories with a small number of samples and enhance the generalization performance of the models.

\begin{table}[t]
    \centering\setlength{\tabcolsep}{4pt}\small
    \caption{Performance (\%) of different methods on test data.}
    \resizebox{\columnwidth}{!}{
    \begin{tabular}{l|c|c|c|ccc}
        \toprule
        Method & Input & Acc  & Edit & F1@10  & F1@20  & F1@50 \\ \midrule
        P4Transformer \cite{fan2021point} & Point & 74.43 &	80.07 &	79.83 &	75.86 &	65.55 \\ 
        PPTr \cite{wen2022point} & Point & 77.46 &	80.15 &	 81.72 & 78.56 &	69.50 \\ \midrule
        C3D \cite{tran2015learning} & Depth & 80.34 &	73.64 &	77.53 &	75.20 &	68.25 \\ 
        R(2+1)D \cite{tran2018closer} & Depth & 80.69 & 74.09 & 74.09	& 75.56	 & 68.20 \\   
        SlowFast \cite{feichtenhofer2019slowfast} &  Depth & 80.79	& 78.28 &	76.86	& 75.24	& 68.70 \\    \midrule
        DPMix  &	Point + Depth  & \bf{85.11} & \bf{88.07} &	\bf{89.91} &	\bf{88.32} & 	\bf{82.76}\\    
        \bottomrule
    \end{tabular}
    }
    \label{tab:test_results}
\end{table}

As illustrated in Fig.~\ref{fig:confusion}, we further count misclassified action categories, hoping to provide inspiration for subsequent improvement methods.
The statistical results of Confusion matrix show that actions between adjacent frames in a video are often misclassified.
In addition, categories with similar action meanings are often misjudged as each other, such as the action "Rest" and the action "Stop".

\subsection{Model Ensemble}
To leverage the complementary nature of predictions from depth modeling methods and point cloud modeling methods, we employ an ensemble approach to combine the predictions of various models.
Using the average aggregation strategy, we aggregate the action probabilities of each model to formulate the final predictions.

As shown in Table~\ref{tab:test_results}, the submission results of each method on test data reports the following three metrics: frame-wise accuracy (Acc), segmented edit distance, as well as segmented F1 scores at the overlapping thresholds of 10\%, 25\%, and 50\%. Overlapping thresholds are determined by the IoU ratio.


On accuracy metric, the depth video experts outperform the point cloud video experts. However, in terms of the edit distance metric, the point cloud video experts achieve better performance.
The possible reason could be that
in the Transformer structure, each point region performs attention with all the other point regions. However, in convolutional-based approaches, the temporal convolutional kernel is typically not very large due to computational constraints, resulting in poor continuity performance in long sequences.
We attempt to increase the temporal convolutional kernel size to extract long-term features, and observe some improvement on the edit metric. However, it still does not surpass the performance of the point cloud method.
The results of our DPMix method, demonstrated in Table~\ref{tab:test_results}, have secured the first position for the total action accuracy in the Action Segmentation Track of the HOI4D Challenge 2023.

\section{Conclusion}
In conclusion, this report has presented our study on HOI4D Dynamic Point Cloud Perception Task Challenge for Action Segmentation 2023.
Our method fuses point cloud video experts and deep video experts, thus enabling better temporal and spatial modeling of long videos.
Although there is still room for improvement in the classification accuracy of some action categories, this work effectively improves the accuracy of most action recognition.
With further performance increase from the model ensemble, our final submission achieves the first rank on the leaderboard in terms of frame-wise action segmentation accuracy.

\section*{Acknowledgments}
This work is supported by the Fundamental Research Funds for the Central Universities (No. 226-2023-00048).

{\small
\bibliographystyle{ieee_fullname}
\bibliography{egbib}
}

\end{document}